\documentclass[runningheads]{llncs}
\usepackage{graphicx}
\usepackage{color}  
\usepackage{times}  
\usepackage{helvet}  
\usepackage{courier}  
\usepackage[hyphens]{url}  
\usepackage{graphicx} 
\usepackage{natbib}  
\usepackage{caption} 
\DeclareCaptionStyle{ruled}{labelfont=normalfont,labelsep=colon,strut=off} 
\usepackage{graphicx}  
\usepackage{subfigure}
\usepackage{algorithm}
\usepackage{algorithmic}
\usepackage{hyperref}
\usepackage{algorithm}
\usepackage{booktabs}
\usepackage{amsmath}
\usepackage{amssymb}
\usepackage{amsfonts} 
\begin{document}
\title{Motif-aware group signed temporal GCN for anomaly detection in bitcoin trust networks}
%
%
\author{Song LI\inst{1} \and
Jiandong Zhou\inst{2} \and
Chong MO\inst{3} \and
Jin LI\inst{3} \and
Kwok Fai TSO \inst{4} \and
Yuxing Tian \inst{5}}
\authorrunning{S. Author et al.}
%
\institute{ International Digital Economy Academy(Futian) \email{sli228-c@my.cityu.edu.hk} \and
University of Oxford \email{jiandzhou3-c@my.cityu.edu.hk} \and Harbin Institute of Technology, ShenZhen \email{1940072595@qq.com} \and Xi’an Jiaotong University \email{jinlimis@xjtu.edu.cn} \and City University of Hong Kong \email{msgtso@cityu.edu.hk} \and Xi'dian University \email{1919035735@qq.com}
}
\maketitle              
\begin{abstract}
Graph convolutional networks (GCNs) is a class of artificial neural networks for processing data that can be represented as graphs. In this study, we focus on link classification on signed temporal networks, where each edge has a sign and a timestamp. We consider the evolving nature and incorporate both local and global information of the network. More specifically, a motif matrix is computed at each snapshot and used in the GCN aggregation process to capture the local topological information within each snapshot. Group balance theory is used to incorporate global information across time. The ultimate node embeddings at each timestamp are computed as the concatenation of local and global embeddings, which are computed by a fixed-window moving average to incorporate temporal information. Experimental results on bitcoin-alpha and bitcoin-otc datasets show that the proposed model outperforms those in the literature.

\keywords{GCN \and temporal \and Motif \and Anomaly detection \and Signed graph}
\end{abstract}

\section{Introduction}
Research on anomaly detection can date back to the 1980s. Since most data in the real world is relational, graph anomaly detection - identifying anomalous graph objects (nodes, edges and so on) in graphs - has been an important data mining paradigm since the beginning. In recent years, due to the booming growth of finance, e-commerce and security need, graph anomaly detection has been receiving increasing interest. In the past, graph anomaly detection relies heavily on human experts' domain knowledge. Lately, the development of machine learning and deep learning technologies has greatly saved human labor and increased the accuracy and efficiency of identifying potential anomalies.

In the literature, the bitcoin trust network - a who-trust-whom network of people who trade using Bitcoin platforms called Bitcoin Alpha and Bitcoin OTC \cite{kumar2016edge,kumar2018rev2} - has been widely studied. Members in the network rate other members in the scale of $-10$ (total distrust) to $10$ (total trust), which can be a good measure to prevent fraudulent and risky users. Existing literature mainly falls into two categories: the traditional methods and deep-learning based methods. For deep-learning methods, GCN-based models are mostly used. For example, \cite{grassia2021wsgat} proposed wsGAT, an extension of the Graph Attention Network to handle graphs with signed and weighted links. \cite{pareja2020evolvegcn} proposed EvolveGCN which uses RNN to update the GCN parameters. \cite{liu2021motif} proposed MTSN, a dynamic network embedding framework that simultaneously models the local structures (motifs) and temporal evolution for dynamic attributed networks. \cite{raghavendra2022signed} proposed a temporal GCN model that utilizes balance theory to guide the training process. \cite{wu2020hierarchical} proposed a hierarchical attention signed network (HASN) to incorporate motif as well as balance information. According to \cite{harary1953notion}, balance theory can not fully model the structure of real signed graphs, thus \cite{liu2021signed} assumes that a signed graph has multiple latent groups and proposed the group signed graph neural network(GS-GCN) to incorporate both global and local information. 
Though many studies have been conducted, few of them simultaneously considers the temporal, local structural and group balance information within the network. In this paper, we propose a deep-learning based temporal graph anomaly detection model to detect fraud in the cryptocurrency trust network. Our contributions are as follows:
\begin{itemize}
    \item We propose the Motif Group-Signed Temporal Graph Convolutional Network (MGS-TGCN), which is a discrete-time temporal GCN model that simultaneous considers the temporal, local structural and group balance information in the signed network;
    \item A motif matrix (each element represents the number of motifs appearing on the edge) is computed and used to replace the adjacency matrix in the GCN aggregation process to incorporate local topological information;
    \item We incorporate global information by the concatenation of global node embeddings. They are computed by the multiplication of node-group attentions and the cross-time global embeddings.
\end{itemize}

\section{Related work}
\subsection{Temporal graph neural network}
Graph convolutional network (GCN) is a class of artificial neural networks for processing data that can be represented as graphs GCN\cite{kipf2016semi}.  Many variants of GCN have been proposed, such as GraphSAGE\cite{hamilton2017inductive} and GAT\cite{velivckovic2017graph}. In this study, we use GCN as the backbone. A 2-layer GCN has the forward equation of the following format.
\begin{equation}
    Z = f(X,A) = \text{softmax}(\hat{A}\,\sigma(\hat{A}XW^{(0)})W^{(1)})
\end{equation}
where $\sigma$ is the activation function, $\sigma(\hat{A}XW^{(0)})$ is the outcome of the first layer, $\hat{A} = \Tilde{D}^{-\frac{1}{2}}\Tilde{A}\Tilde{D}^{-\frac{1}{2}}$ is the scaled adjacency matrix, $\Tilde{D}$ and $\Tilde{A}$ are the degree matrix and adjacency matrix with self-loops, $X$ is the attributes matrix, $W^{(0)}$ and $W^{(1)}$ are the weight matrices.
Despite the plethora of different GCN models on graphs, they cannot handle the real-world networks which are dynamic in nature, where features or connectivity are evolving over time. Some temporal GCN models are proposed to learn from dynamic graphs and are proven to outperform the static approaches. For discrete-time temporal GCN models, the graph data is firstly merged into snapshots. The prediction result at each snapshot is achieved by modeling on previous snapshots \cite{pareja2020evolvegcn,sankar2020dysat}. An illustration of discrete-time temporal graph is shown in Figure 1. The study in this paper falls into the scope of discrete-time temporal GCN. 

\begin{figure}[h]
\centering
\includegraphics[width=3.5in]{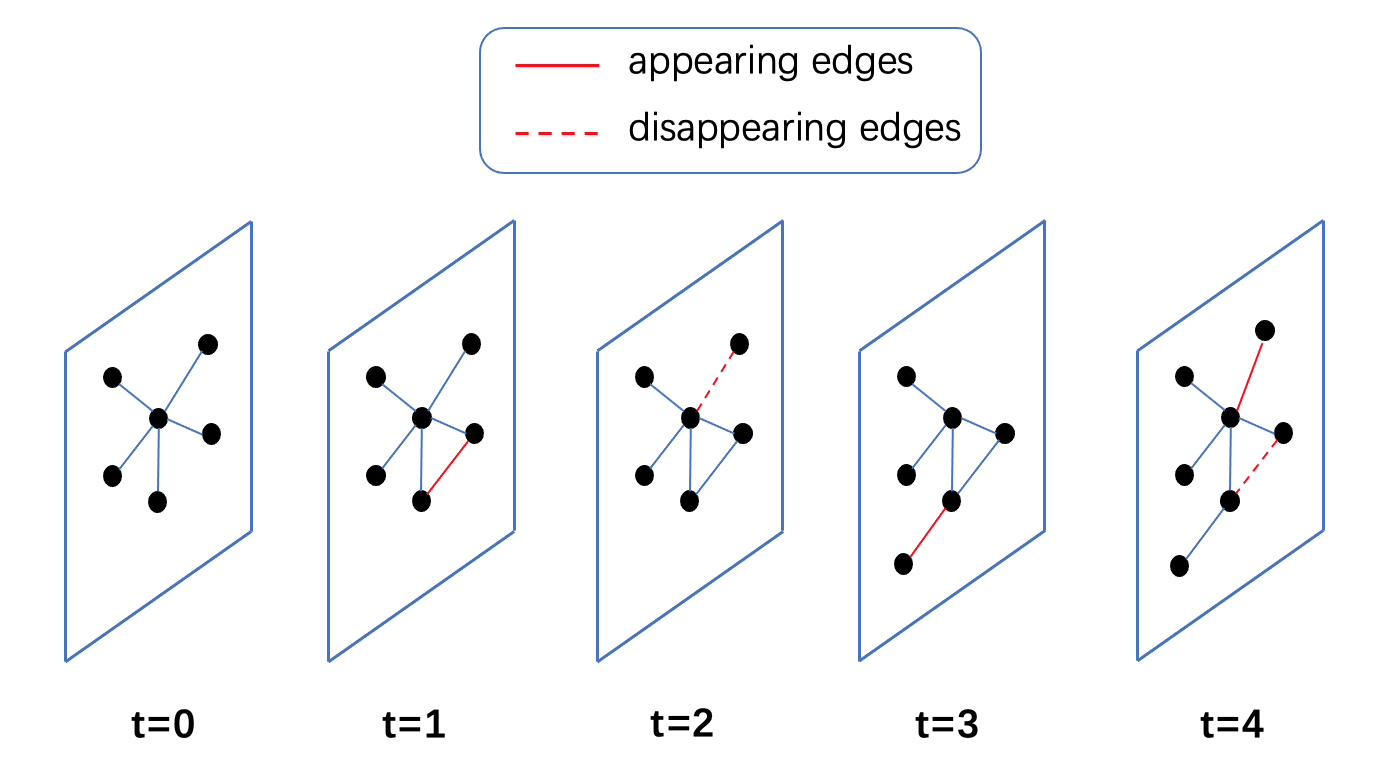}
\caption{An illustration of discrete-time temporal graph.}
\label{fig:temporal_graph}
\end{figure}

\subsection{Balance theory}
Balance theory, proposed by Fritz Heider, conceptualizes the cognitive consistency motive as a drive toward psychological balance. It states that signed relationships tend to form balanced triads, where there are odd number of positive edges. In this study we use balance theory to differentiate the information flow on a positive edge or a negative edge, which helps differentiate the four relationships (friend-friend / friend-enemy / enemy-friend / enemy-enemy). 
Balance theory is shown equivalent to the assumption that nodes can be divided into two conflicting groups (\cite{harary1953notion}), which is too ideal for signed graphs in the real world. Thus, existing methods which simply use the balance theory can not model the true underlying structure of real signed graphs. To overcome this limitation, we use a similar but simpler latent group approach as in \cite{liu2021signed} in this study. An illustration of balanced triads and unbalanced triads is shown in Figure 2. For balanced triads, there are odd number of positive signs.

\begin{figure}[h]
\centering
\includegraphics[width=3.5in]{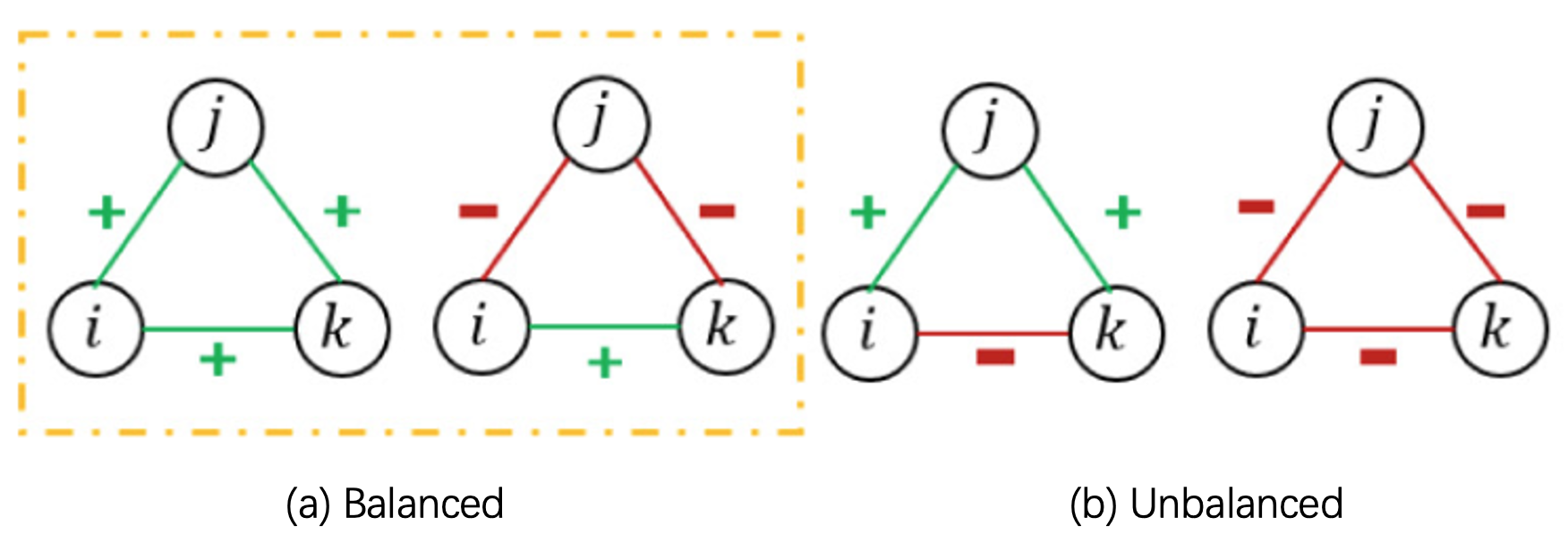}
\caption{Balanced triads and unbalanced triads.}
\label{fig:balance_theory}
\end{figure}

\subsection{Motif detection}
Motifs are fundamental patterns in networks that could be regarded as network blocks. Motif discovery has been widely applied in various scientific problems, such as subgraph mining. In the literature, many motif discovery algorithms, such as MFinder\cite{kashtan2004efficient}, Grochow\cite{grochow2007network}, MODA\cite{omidi2009moda}, PGD\cite{ahmed2015efficient} have been proposed. In this study, \href{https://github.com/nkahmed/PGD}{PGD}(a parallel parameterized graphlet decomposition library) is used to get the number of motifs for each edge. The motif types used in this paper are triangle, 2-star, 4-clique, 4-chordalcycle, 4-tailedtriangle, 4-cycle, 3-star and 4-path, as shown in Figure 3.

\begin{figure}[h]
\centering
\includegraphics[width=3.5in]{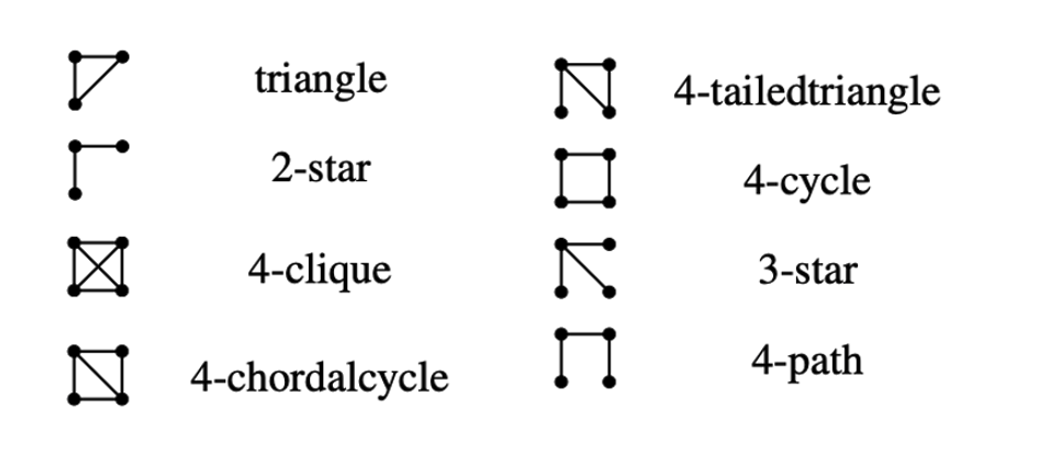}
\caption{The types of motif used in this paper.}
\label{fig:motif_types}
\end{figure}

\section{Methods}
\subsection{Notations}
Throughout this paper, we use subscript $t$ to denote the time index. We assume all graphs have $N$ nodes, even though it will change at different snapshots. At time $t$, the nodes and edges of the graph can be denoted as $(V_{t}, E_{t})$, the input data consists of the pair $(A_{t} \in \mathcal{R}^{n\times n}, X_{t} \in \mathcal{R}^{n\times d}, M_{t} \in \mathcal{R}^{n\times n}, S_{t} \in \mathcal{R}^{n\times n})$, where $A_{t}$ is the adjacency matrix, $X_{t}$ is the feature matrix, $M_{t}$ is the motif matrix, $S_{t}$ is the sign matrix. Table 1 is a summary of all the notations used in this paper.

\begin{table}
\centering
\caption{Notations.}
\begin{tabular}{cc}
\toprule[1.5pt]
Variable & Definition  \\ \midrule[1pt]
$t$  &  time index  \\
$G_{t}$  &  graph at time $t$ \\
$|V_{t}|$ & number of nodes at time $t$ \\ 
$|E_{t}|$ & number of edges at time $t$ \\ 
$A_{t}$ & adjacency matrix at time $t$ \\ 
$X_{t}$ & feature matrix at time $t$ \\ 
$Y_{t}=(y_{1},...,y_{n})$ & labels of edges at time $t$ \\
$P_{t}$ & predicted softmax probability at time $t$\\ 
$\hat{Y_{t}}$ & predicted labels at time $t$ \\
$APM_{t}$ & assignment matrix at time $t$ \\
$S_{t}$ & sign matrix at time $t$ \\ 
$M_{t,i}$ & motif matrix of type i at time $t$ \\ 
$\alpha_{i}$ & learnable weight on $M_{t,i}$ for all t \\
$M_{t}$ & motif matrix at time $t$ \\ 
$S_{t}^{+}$ & $S_{t}$ with only positive elements \\
$S_{t}^{-}$ & $S_{t}$ with only negative elements \\
$Z_{t,c}$ & $1\times d$ embedding for group $c$ at time $t$ \\
$Q_{v,c}$ & node $v$ and $i_{th}$ group attention at time $t$\\
$W_{1}$ & $C \times C$ attention transformation matrix \\
$W_{2}$ & weight for global embedding computation \\
$W_{3}$ & weight for positive aggregation \\
$W_{4}$ & weight for negative aggregation \\
$W_{5}$ & weight for dynamic embeddings computation \\
$Emb_{t,i}$ & embedding of node i at time $t$ \\
$Global\text{-}Emb_{t,i}$ & global embedding of node i at time $t$ \\
$Link\text{-}Emb_{t,(i,j)}$  & embedding of link (i,j) at time $t$ \\
$Graph\text{-}Emb_{t}^{+}$ & positive graph node embedding \\
$Graph\text{-}Emb_{t}^{-}$ & negative graph node embedding \\
$Motif\text{-}Emb_{t}$ & motif embedding at time $t$ \\
$Dy\text{-}Emb_{t}^{+}$ & positive dynamic embedding \\
$Dy\text{-}Emb_{t}^{-}$ & negative dynamic embedding \\
\bottomrule[1.5pt]
\end{tabular}
\end{table}

\subsection{Node representation}
In MGS-TGCN, computation of the ultimate node representations for each snapshot includes computing the following: global embedding $Global\text{-}Emb_{t}$, the static graph embedding $Graph\text{-}Emb^{+}_{t}$ /  $Graph\text{-}Emb^{-}_{t}$, the static motif embedding $Motif\text{-}Emb_{t}$, the dynamic embedding $Dy\text{-}Emb_{t}^{+}$, $Dy\text{-}Emb_{t}^{-}$. In this section, we show details of the computation of these embeddings.

\subsubsection{Global node representation}
The global node representations for each snapshot are computed by the following: initialize cross-time global embeddings, initialize node-group attention values, temporal assignment probability matrix computation, global embedding computation.
\begin{itemize}
    \item \textbf{Global embeddings initialization}: initialize global embeddings $Z_{ini}=\{ {Z_{ini,1},...,Z_{ini,C}} \}$, where $Z_{ini,c} \in \mathbb{R}^{V\times d}$;
    
    \item \textbf{Node-group attention values initialization}: The node embeddings are computed by the transformation of the input node attributes $\mathbf{X}$ through a multi-layer perceptron (MLP), i.e.:
    \begin{equation}
        \mathbf{X^{'}} = MLP(\mathbf{X}). \forall v \in \mathcal{V}
    \end{equation}
    where $\mathbf{X^{'}} \in \mathbb{R}^{d}$. The assignment probability is a softmax normalization of the attention values between nodes representations and group embeddings. For example, the attention value between node $v$ and group $c$ is:
    \begin{equation}
        Q_{v,c}^{(0)} = Z_{ini,c}\mathbf{X_{v}'}^{T}
    \end{equation}
    To incorporate graph structural information, we conduct two-layer signed aggregation on $Q^{\mathcal{V}\times C}$ to compute the initial attention values, which follows the following equation:
    \begin{align}
    \begin{split}
    & Q'^{(0)}  = 
    \\ & (A^{+}\sigma(A^{+}Q^{(0)}W^{(0)})W^{(1)}) || (A^{-}\sigma(A^{-}Q^{(0)}W^{(0)})W^{(1)})
    \end{split}
    \end{align}
    where  $A^{+} = S^{+} \odot A$, $A^{-} = S^{-} \odot A$, $S^{+}$ and $S^{-}$ is the cross-time positive and negative sign matrix

    \item \textbf{Temporal assignment probability matrix computation}: $Q'$ is computed in the cross-time full graph context. The assignment probability matrix of each snapshot needs to be adjusted to account for the structural difference. In this study, we use a shared weight matrix $W_{4}$ and the adjacency matrix $A_{t}$ to transform the attention values for time $t$, i.e.:
    \begin{equation}
        Q'_{t} = A_{t} \cdot Q'^{(0)} \cdot W_{1}
    \end{equation}

    The assignment probability matrix at time t is computed by:
    \begin{equation}
        APM_{v}^{t} = softmax(Q'_{t,v}) = \frac{exp(Q'_{t,v,c})}{\sum_{c=1}^{C}exp(Q'_{t,v,c})}
    \end{equation}

    \item \textbf{global embedding computation}: The node global embeddings at time $t$ is computed as the average within a time-window. Here we set the window-size to be two.
    \begin{equation}
        Z_{t} = W_{2}*APM_{v}^{t} \cdot Z_{ini} + (1-W_{2})*Z_{t-1}
    \end{equation}
\end{itemize}

\subsubsection{Graph embedding}
The positive and negative graph embedding ($Graph\text{-}Emb^{+}_{t}$ / $Graph\text{-}Emb^{-}_{t}$) are generated by the neighborhood aggregation process as in equation (1), which can capture the graph structural information within each snapshot. According to the balance theory, the information coming from positively linked nodes and negatively linked nodes should be different. Therefore, we use different strategy to aggregate information through edges with different signs. Details of the generation process are shown as following.
\begin{equation}
\begin{split}
    Graph\text{-}Emb^{+}_{t} = ReLU((S_{t}^{+} \odot A_{t}) \cdot Dy\text{-}Emb_{t-1}^{+} \cdot W_{3} \\ 
    + (S_{t}^{-} \odot A_{t}) \cdot Dy\text{-}Emb_{t-1}^{-} \cdot W_{4} + b^{+})
\end{split}
\end{equation}
\begin{equation}
\begin{split}
    Graph\text{-}Emb^{-}_{t} = ReLU((S_{t}^{-} \odot A_{t}) \cdot Dy\text{-}Emb_{t-1}^{-} \cdot W_{3} \\
    + (S_{t}^{+} \odot A_{t}) \cdot Dy\text{-}Emb_{t-1}^{+} \cdot W_{4} + b^{-})
\end{split}
\end{equation}

\subsubsection{Motif embedding}
The motif embedding $Motif\text{-}Emb_{t}$ is also computed as in equation (1) with the adjacency matrix $A_{t}$ replaced by the motif matrix $M_{t}$. Intuitively, the number of motifs is used as the weight to aggregate information from the neighbours. The motif-counting library \href{https://github.com/nkahmed/PGD}{PGD} is used to compute the eight types of motifs. Details of the motif embedding generation process with two-layers GCN are shown as follows.
\begin{equation}
    M_{t} = \sum_{i=1}^{8}\alpha_{i}M_{t,i}
\end{equation}
\begin{equation}
    Motif\text{-}Emb_{t} = \sigma(M_{t}\,\sigma(M_{t}XW^{(0)})W^{(1)})
\end{equation}

\subsubsection{Dynamic embedding}
The positive and negative dynamic node embeddings are computed by the weighted average of history dynamic embeddings within a time window, where the weights are learnable parameters. Motif embeddings are added to the positive and negative graph node embeddings to get the dynamic embeddings at current snapshot. Details are shown as follows.
\begin{equation}
\begin{split}
    Dy\text{-}Emb_{t}^{+} \longleftarrow W_{5} * (Graph\text{-}Emb^{+}_{t} + \beta Motif\text{-}Emb_{t}) \\
    + (1 - W_{5}) * (Dy\text{-}E_{t-1}^{+}) \\
\end{split}
\end{equation}

\begin{equation}
\begin{split}
    Dy\text{-}Emb_{t}^{-} \longleftarrow W_{5} * (Graph\text{-}Emb^{-}_{t} + \beta Motif\text{-}Emb_{t}) \\
    + (1 - W_{5}) * (Dy\text{-}E_{t-1}^{-}) \\
\end{split}
\end{equation}
The parameter $\beta$ is a predefined hyper-parameter. In practice, we set $\beta \in [0.3,0.4]$. $W_{t}$ is a learnable parameter.

\subsubsection{Final node embedding}
The final node embedding at time $t$ is the concatenation of positive dynamic embedding, negative dynamic embedding and global embedding, i.e.
\begin{equation}
Emb_{t} = (Dy\text{-}Emb_{t}^{+} || Dy\text{-}Emb_{t}^{-}||Z_{t})
\end{equation}
In this way, $Emb_{t}$ contains structural, motif and the cross-time global balance information. It is more informative compared with the original GCN embedding.

In our task, the information of $G_{1:t}$ is used to predict the labels $Y_{t}$ of edges at time $t$. From time $1$ to $t-1$, the sign matrix $S_{t}$ is generated by the signs of the edges. Since the sign information at time $t$ is unavailable, we make extreme assumption by setting $S_{t}^{+}$ and $S_{t}^{-}$ to be all-one matrix. In this way, the information from all the neighbors is aggregated in the last timestamp to generate $Graph\text{-}Emb^{+}_{t}$ and $Graph\text{-}Emb^{-}_{t}$.
The embeddings generated by equation (14) of two connected nodes are concatenated to form the edge embedding. Softmax function is used to normalize the embedding to the probability distribution over predicted output classes. Cross-entropy loss is used for back-propogation.

\begin{center}
    \begin{figure*}[h]
    \includegraphics[width=5in]{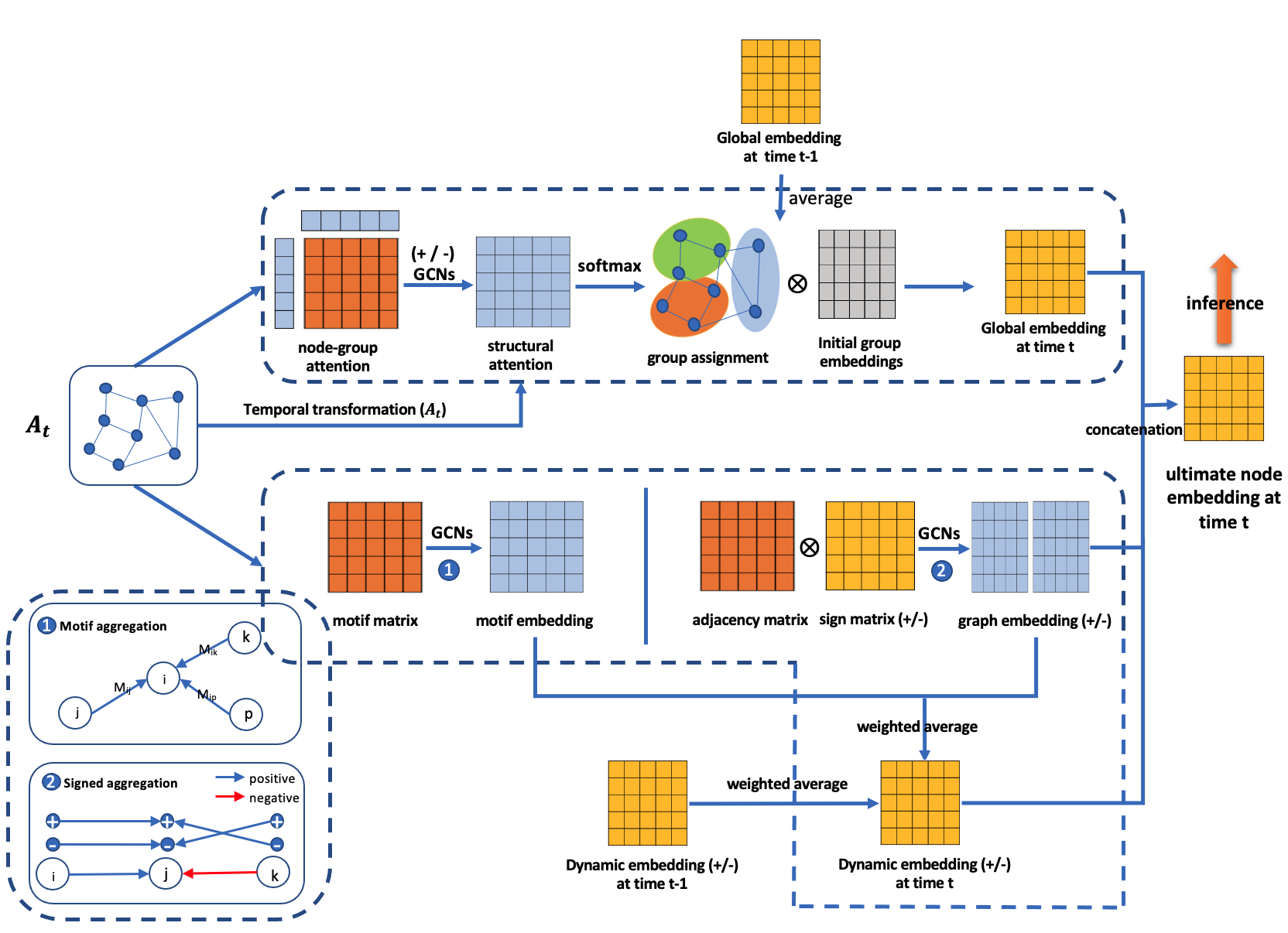}
    \caption{Motif-aware group signed temporal GCN}
    \label{fig:model_architecture}
    \end{figure*}
\end{center}

\section{Experiments}
\subsection{Data description}
The bitcoin-alpha and bitcoin-otc dataset used in this paper are who-trusts-whom networks of people who trade using Bitcoin on the platforms. Bitcoin-alpha has 3777 nodes with 24173 edges, bitcoin-otc has 5881 nodes with 35588 edges. All datasets are divided into training, validation, and testing sets with a proportion of 70\%, 15\%, and 15\% following chronological order. 

\subsection{Baselines}
Three models for static and temporal prediction are used for comparison. The static model is vanilla GCN, the temporal models are GCN-GRU and EvolveGCN \cite{pareja2020evolvegcn}. These models are widely used in the literature as baseline models.

\subsection{Experimental Settings}
All experiments are evaluated on NVIDIA A100 80GB PCle GPU. Adam is adopted for parameter optimization with the learning rate of 0.005 and weight decay of 5e-5. The dimension of node embedding is 200. The number of GCN layers is 2 and the dropout rate for each layer is 0.2. For each experiment, we set the number of warm-up epochs to be 50, the total number of epochs to be 200. In table 2 and table 3, the result of each experiment is the average of the top 5 results from epoch 50 to 200 to avoid random effects. The results shown in this paper are computed as an average of the results of repeatedly conducting the experiment for five times with same experimental settings. 

\subsection{Results and discussion}
The results are shown in Table 2. Since F1 is a better measure for imbalance datasets, F1 score is used as the main metric. It is shown that MGS-TGCN achieves best F1, precision and recall scores on both datasets. To verify whether motif, group balance and signed aggregation in MGS-TGCN has positive effect to the final result, ablation studies are conducted with the same experimental settings. We start from GCN and step-by-step adds motif embedding, replaces the signed aggregation in graph embedding with regular aggregation and adds the global embedding. Results are shown in Table 3, where there is an increasing trend of the scores. This demonstrates the positive effects of these modules.

\begin{table}
\centering
\caption{Bitcoin datasets.}
\begin{tabular}{ccccl}
\toprule[1.5pt]
\multicolumn{1}{l}{} & Nodes & Edges & \begin{tabular}[c]{@{}c@{}}Time steps\\ Train / Valid / Test\end{tabular} & \\ \midrule[1pt]
BC-OTC  & 5881  & 35588 & 8/1/3 &  \\
BC-Alpha & 3777  & 24173 & 8/1/3 & \\ \bottomrule[1.5pt]
\end{tabular}
\end{table}

\begin{table}
\centering
\caption{Expeimental results for edge classificaiton task on bitcoin datasets 
 (upper:BC-Alpha; bottom:BC-OTC)}
\begin{tabular}{lcccc}
\toprule[1.5pt]
 & GCN & GCN-GRU & EvolveGCN & MGS-TGCN \\
\midrule[1pt]
F1 & 0.342 & 0.378 & 0.069 & \textbf{0.433} \\
Acc & 0.683 & \textbf{0.897} & 0.516 & 0.789 \\
Ap & 0.213 & 0.227 & 0.146 & \textbf{0.279} \\
Recall & 0.494 & 0.314 & 0.037 & \textbf{0.484} \\
\bottomrule[0.5pt]
F1 & 0.302 & 0.323 & 0.017 & \textbf{0.342} \\
Acc & 0.807 & \textbf{0.856} & 0.850 & 0.805 \\
Ap & 0.194 & 0.228 & \textbf{0.276} & 0.220 \\
Recall & 0.290 & 0.323 & 0.012 & \textbf{0.387} \\
\bottomrule[1.5pt]
\end{tabular}
\end{table}

\begin{table}
\centering
\caption{The experimental results for ablation studies on bitcoin dataset (upper:BC-Alpha; bottom:BC-OTC)}
\begin{tabular}{lcccc}
\toprule[1.5pt]
 & -Motif & -sign & -global & MGS-TGCN \\
\midrule[1pt]

F1 & 0.312 & 0.334 & 0.360 & \textbf{0.433} \\
Acc & 0.695 & 0.641 & 0.753 & \textbf{0.789} \\
Ap & 0.193 & 0.200 & 0.217 & \textbf{0.279} \\
Recall & 0.624 & \textbf{0.668} & 0.516 & 0.484 \\
\bottomrule[0.5pt]
F1 & 0.283 & 0.319 & 0.319 & \textbf{0.346} \\
Acc & 0.810 & \textbf{0.837} & 0.838 & 0.745 \\
Ap & 0.157 & 0.173 & 0.175 & \textbf{0.211} \\
Recall & 0.385 & 0.394 & 0.410 &  \textbf{0.484} \\
\bottomrule[1.5pt]
\end{tabular}
\end{table}

\subsection{Visual Analysis}
\begin{center}
    \begin{figure}[h]
    \subfigure[Timestamp 6]{
    \includegraphics[width=2.8in]{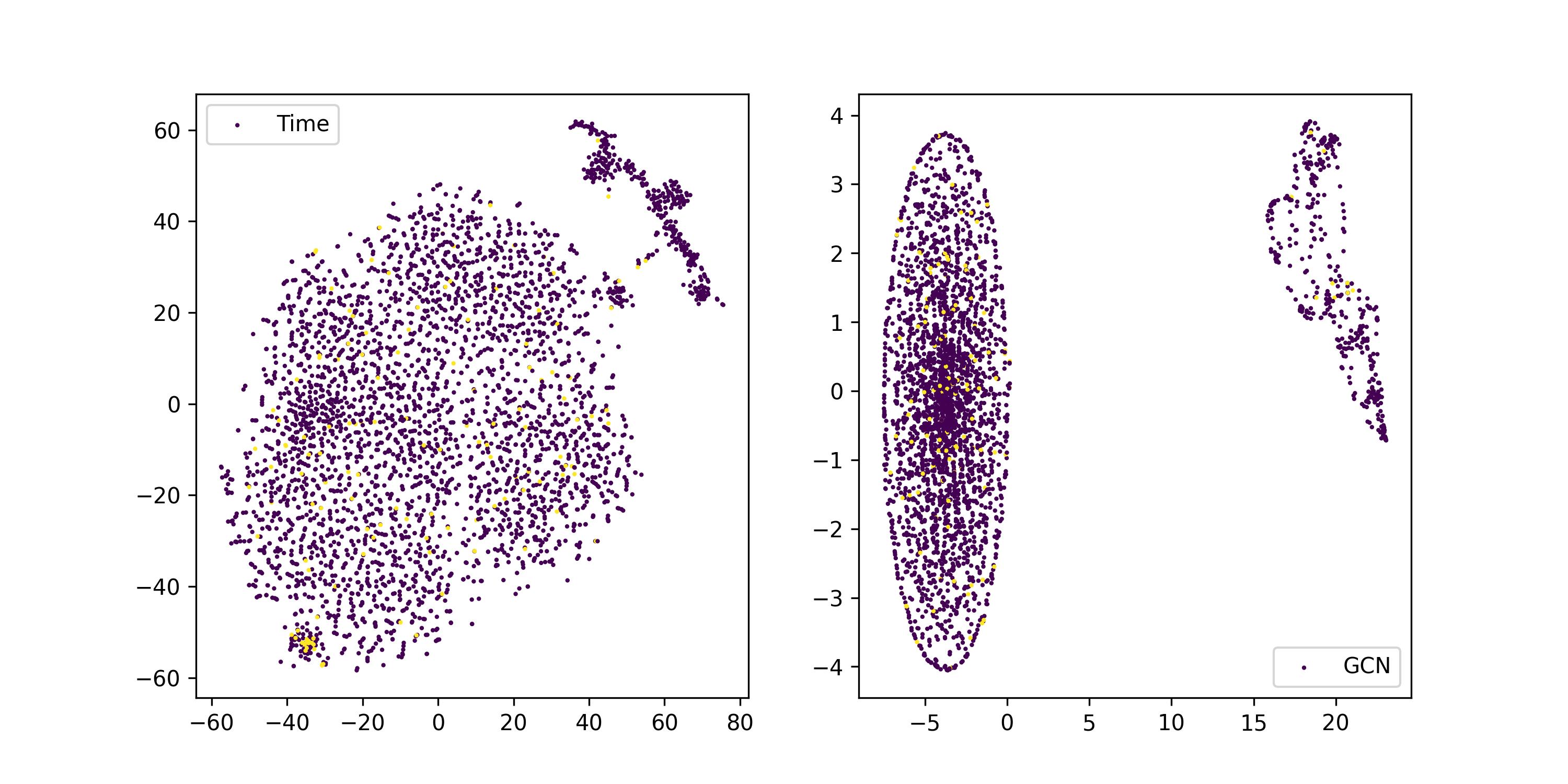}}
    \subfigure[Timestamp 7]{
    \includegraphics[width=2.8in]{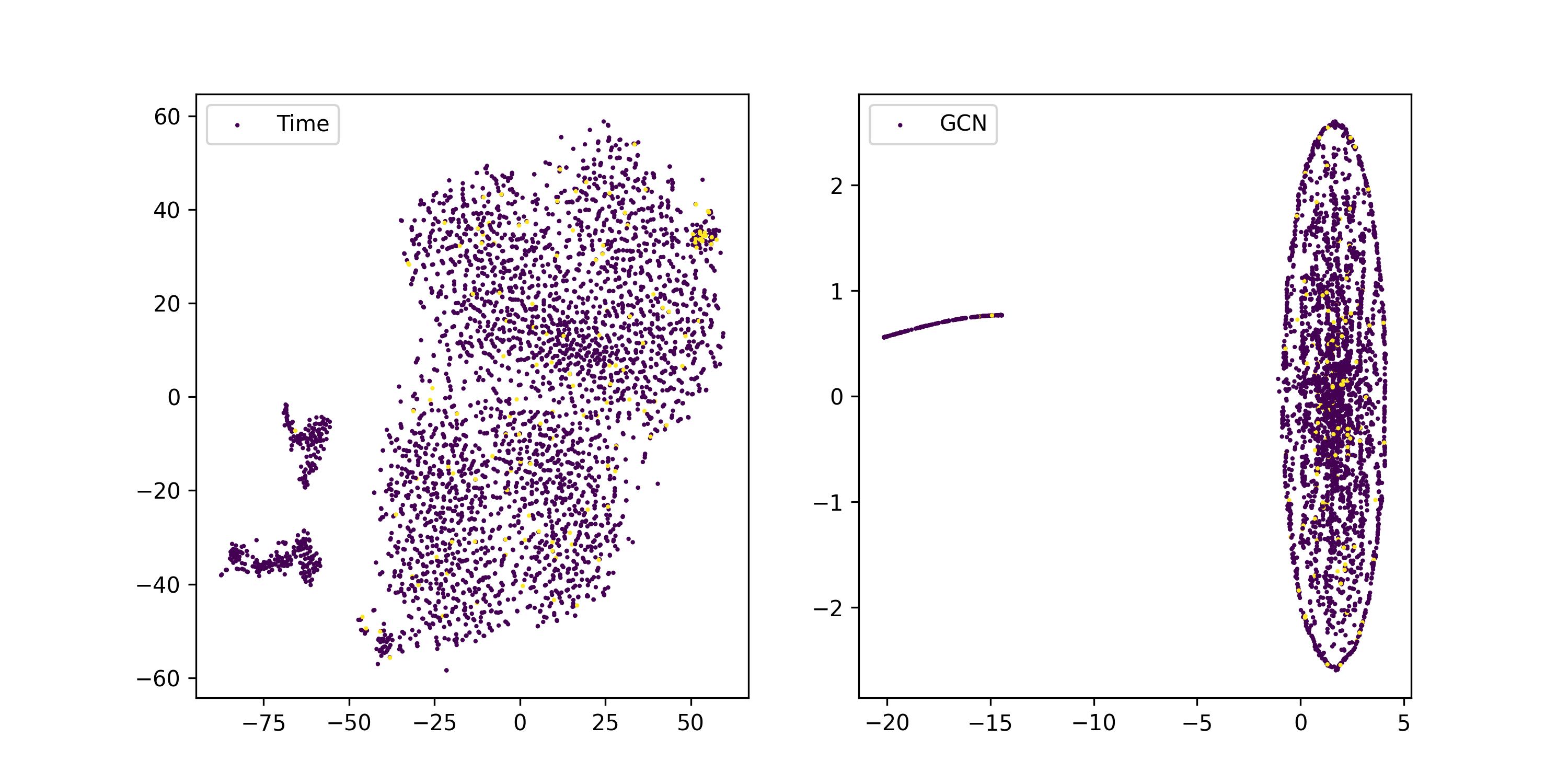}}
    \subfigure[Timestamp 8]{
    \includegraphics[width=2.8in]{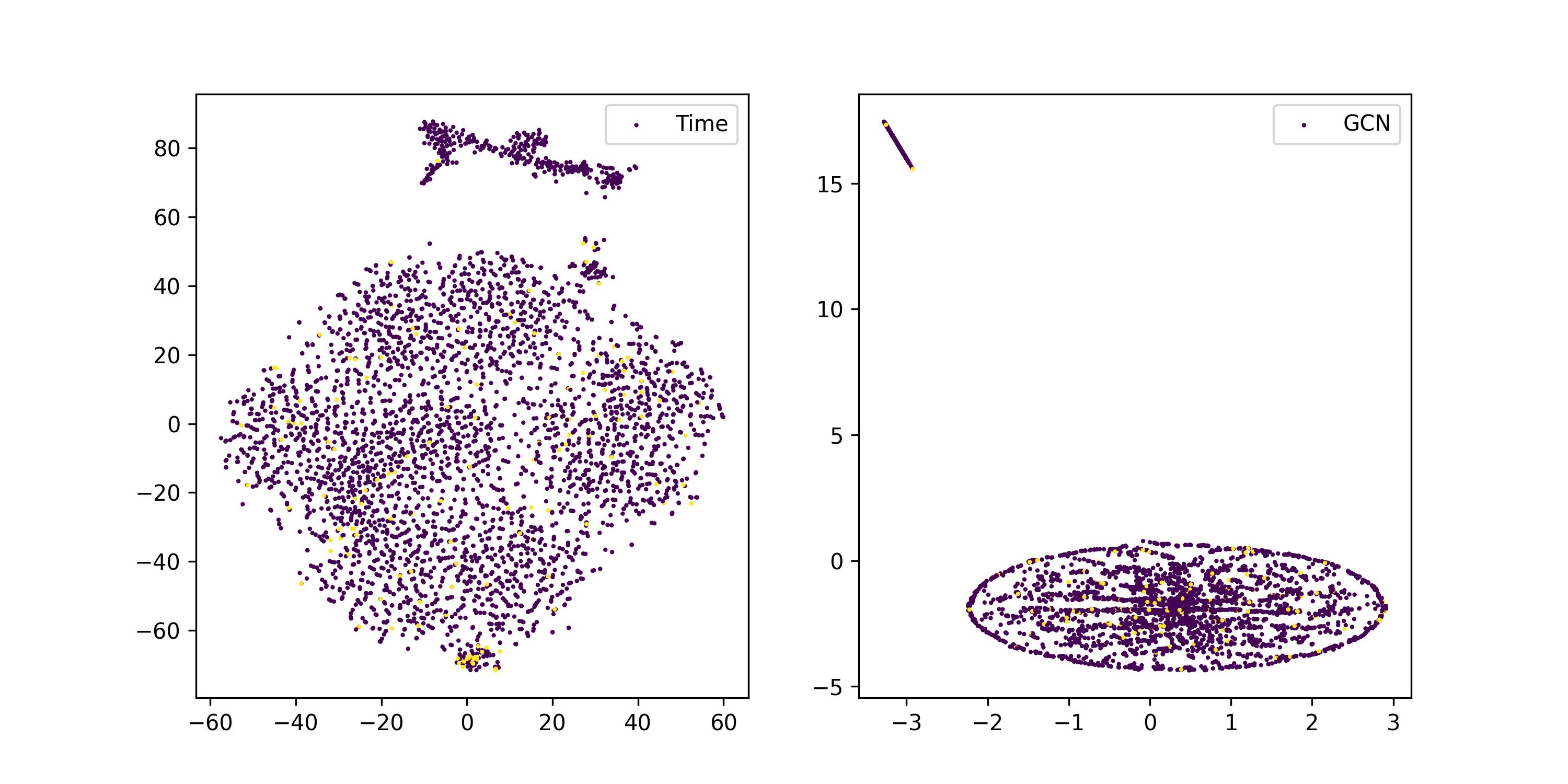}}
    \subfigure[Timestamp 9]{
    \includegraphics[width=2.8in]{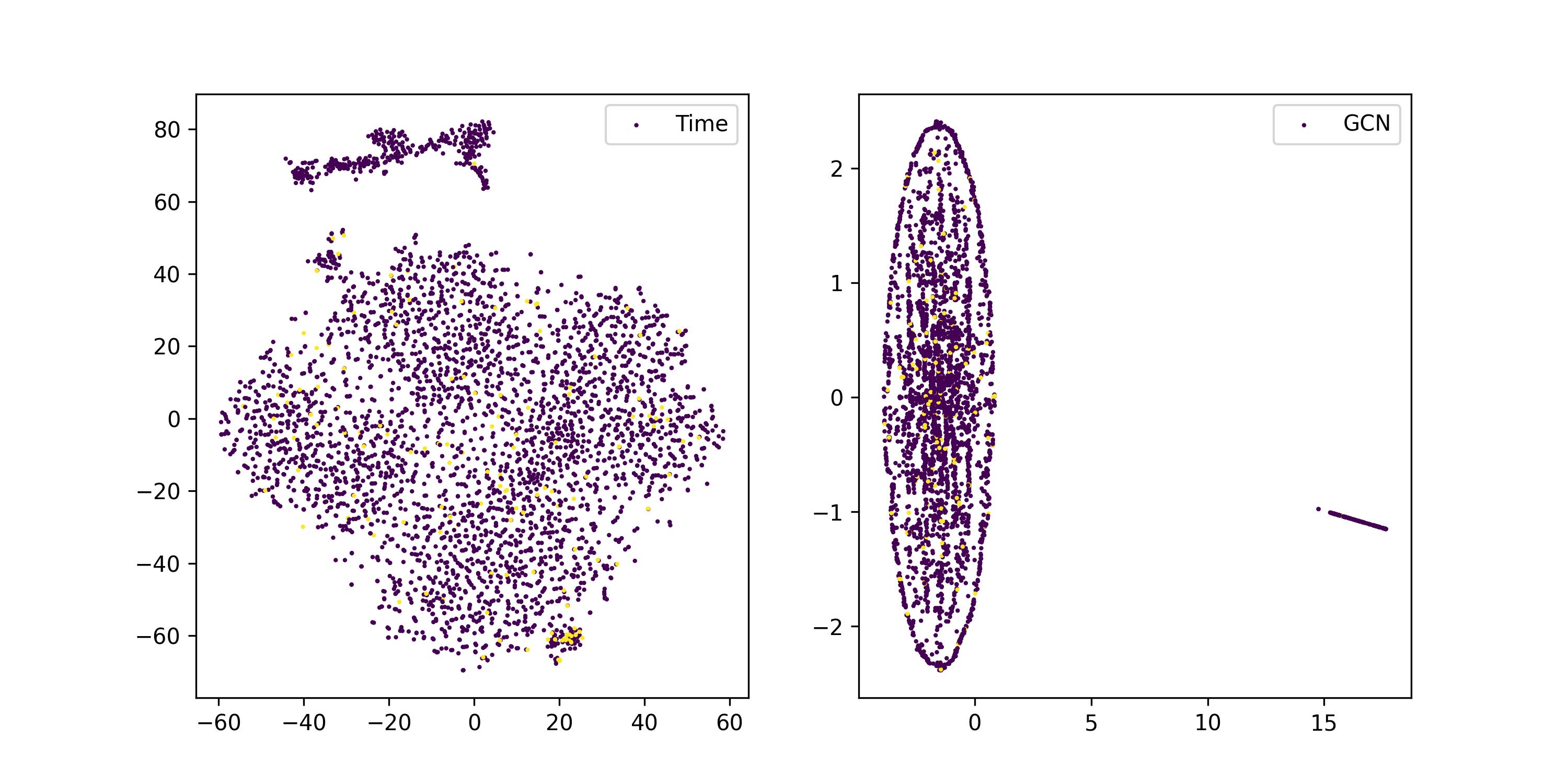}}
    \subfigure[Timestamp 10]{
    \includegraphics[width=2.8in]{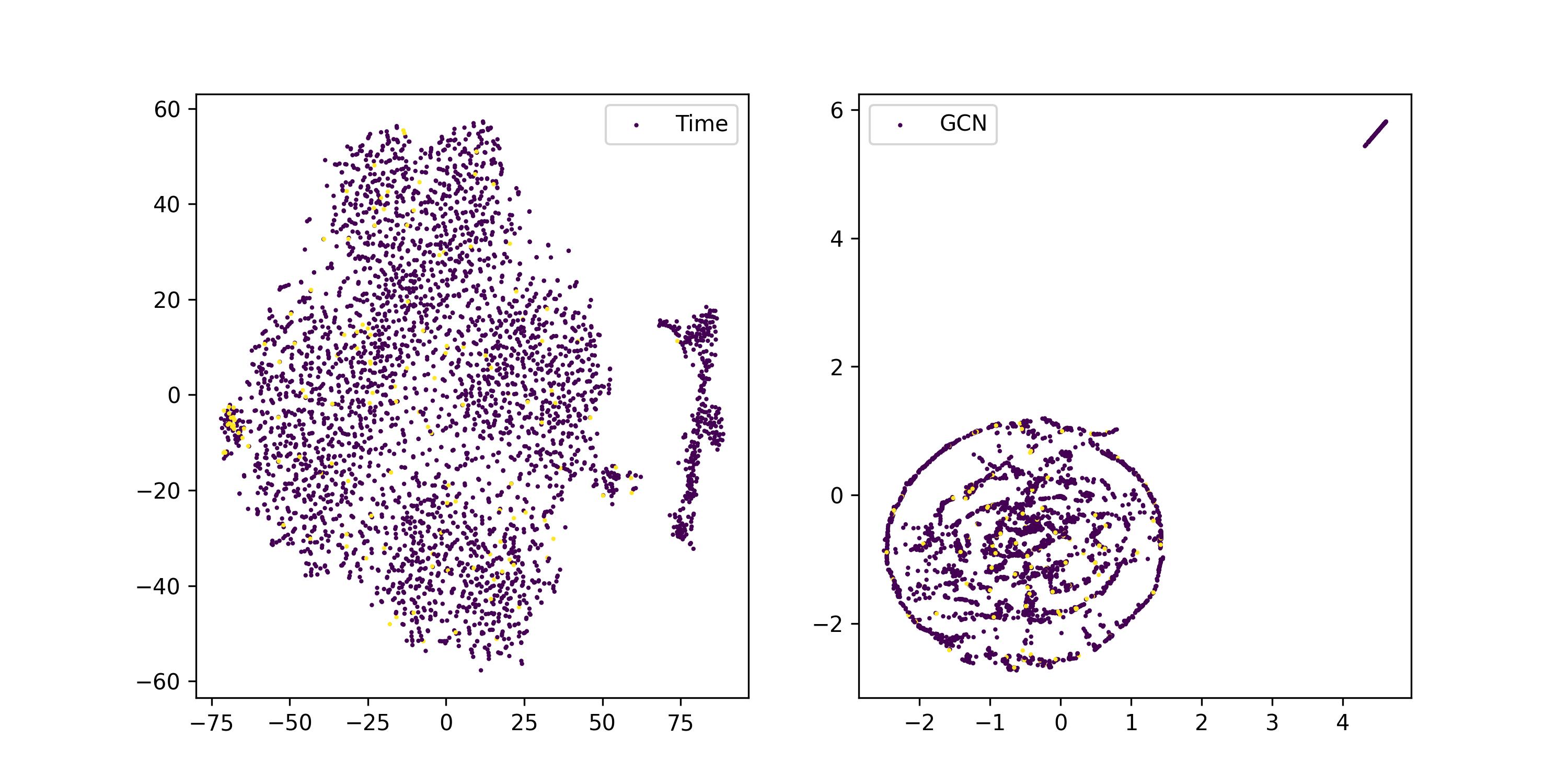}}
    \caption{Dynamic node embedding in bitcoin-alpha dataset }
    \label{fig:alphavis}
    \end{figure}
\end{center}

\begin{center}
    \begin{figure}[h]
    \subfigure[Timestamp 6]{
    \includegraphics[width=2.8in]{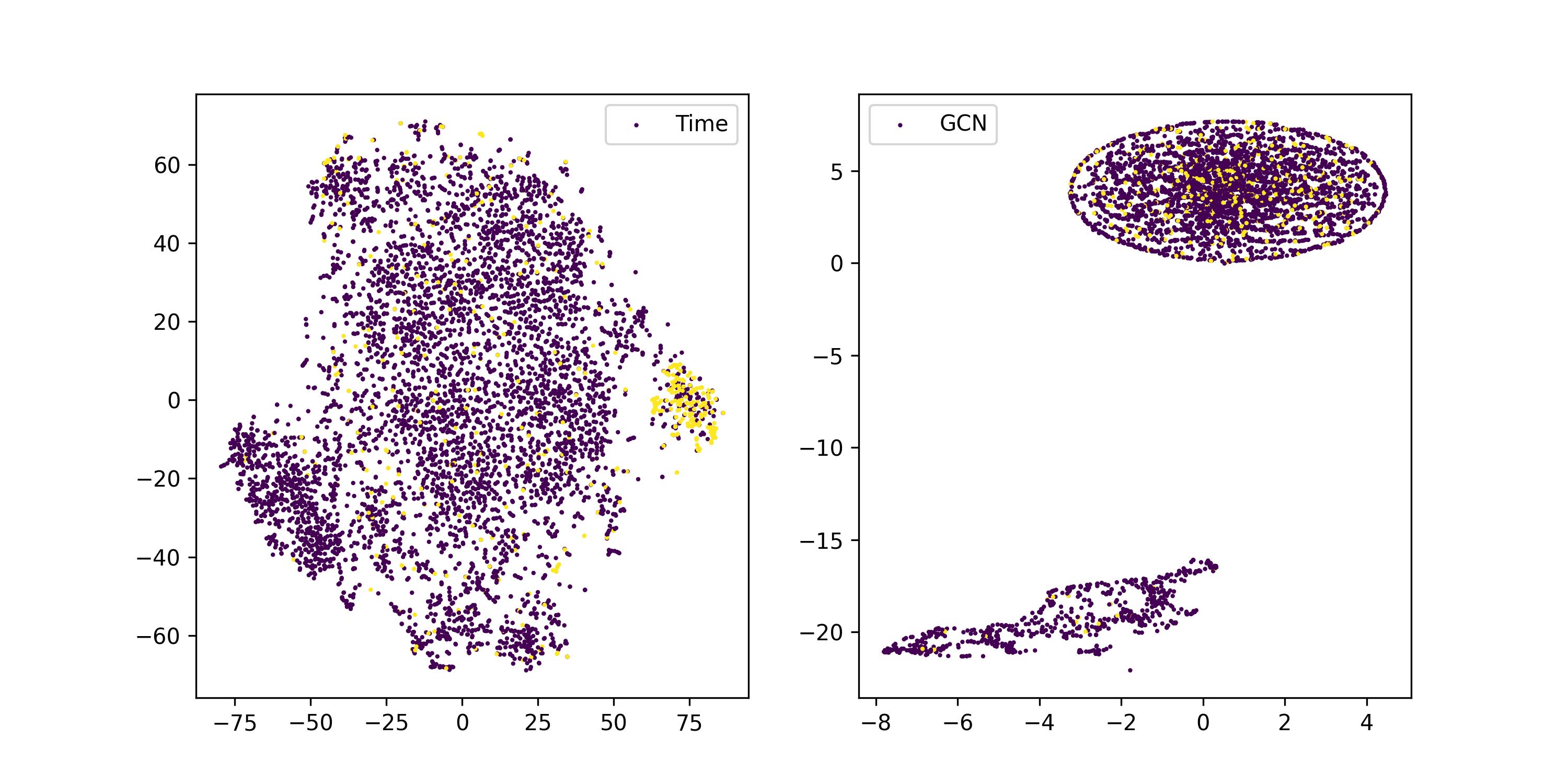}}
    \subfigure[Timestamp 7]{
    \includegraphics[width=2.8in]{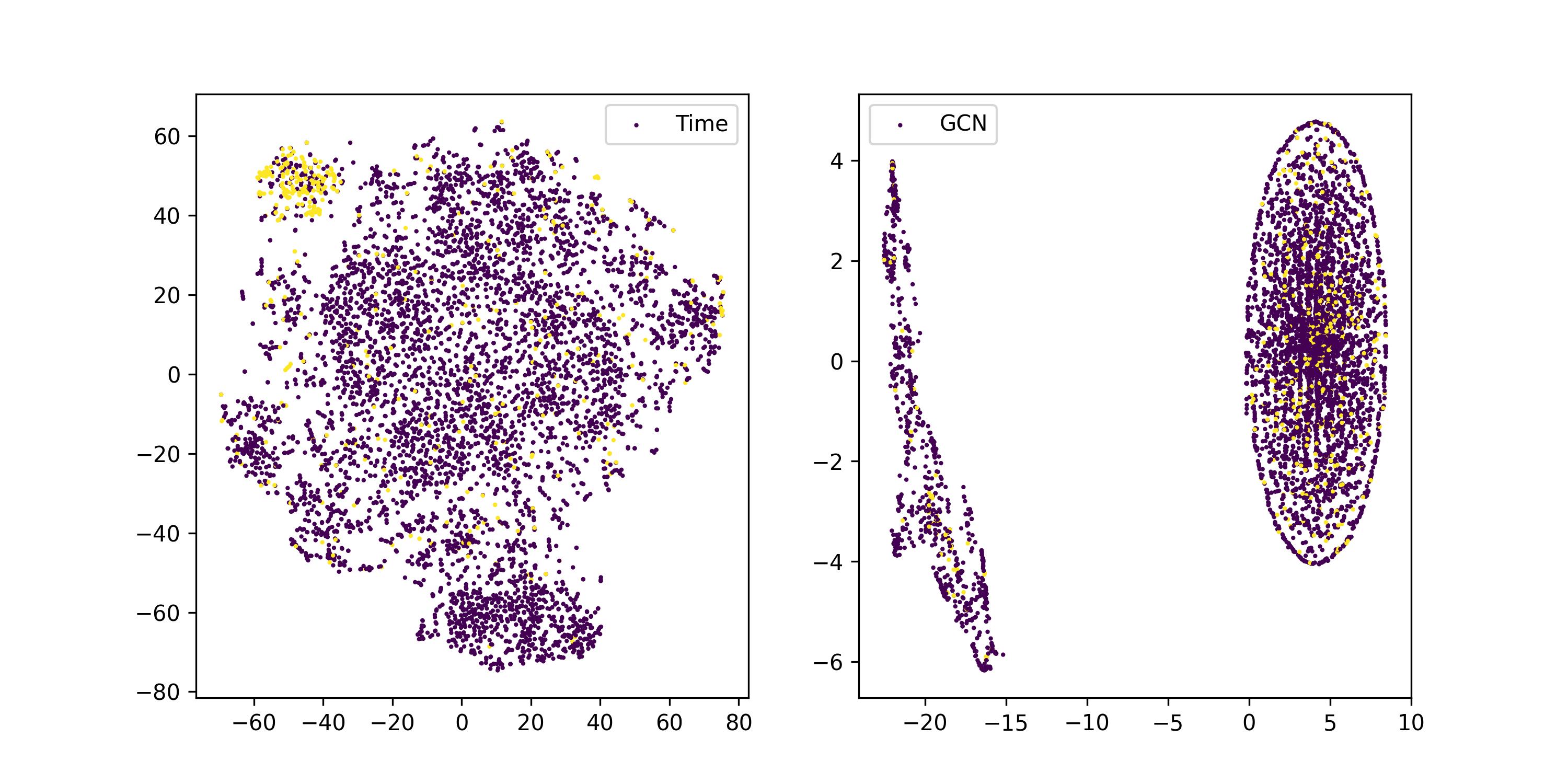}}
    \subfigure[Timestamp 8]{
    \includegraphics[width=2.8in]{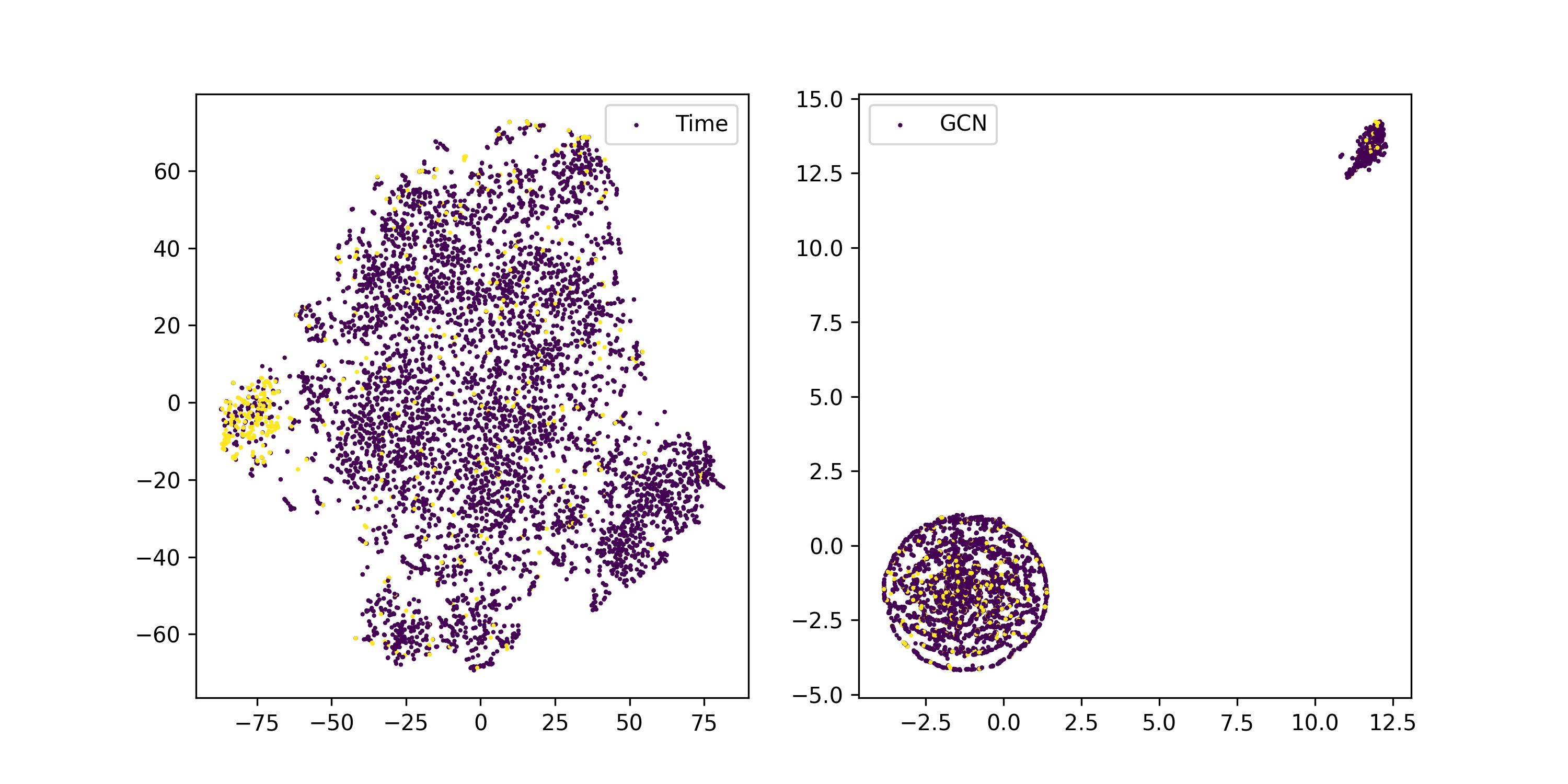}}
    \subfigure[Timestamp 9]{
    \includegraphics[width=2.8in]{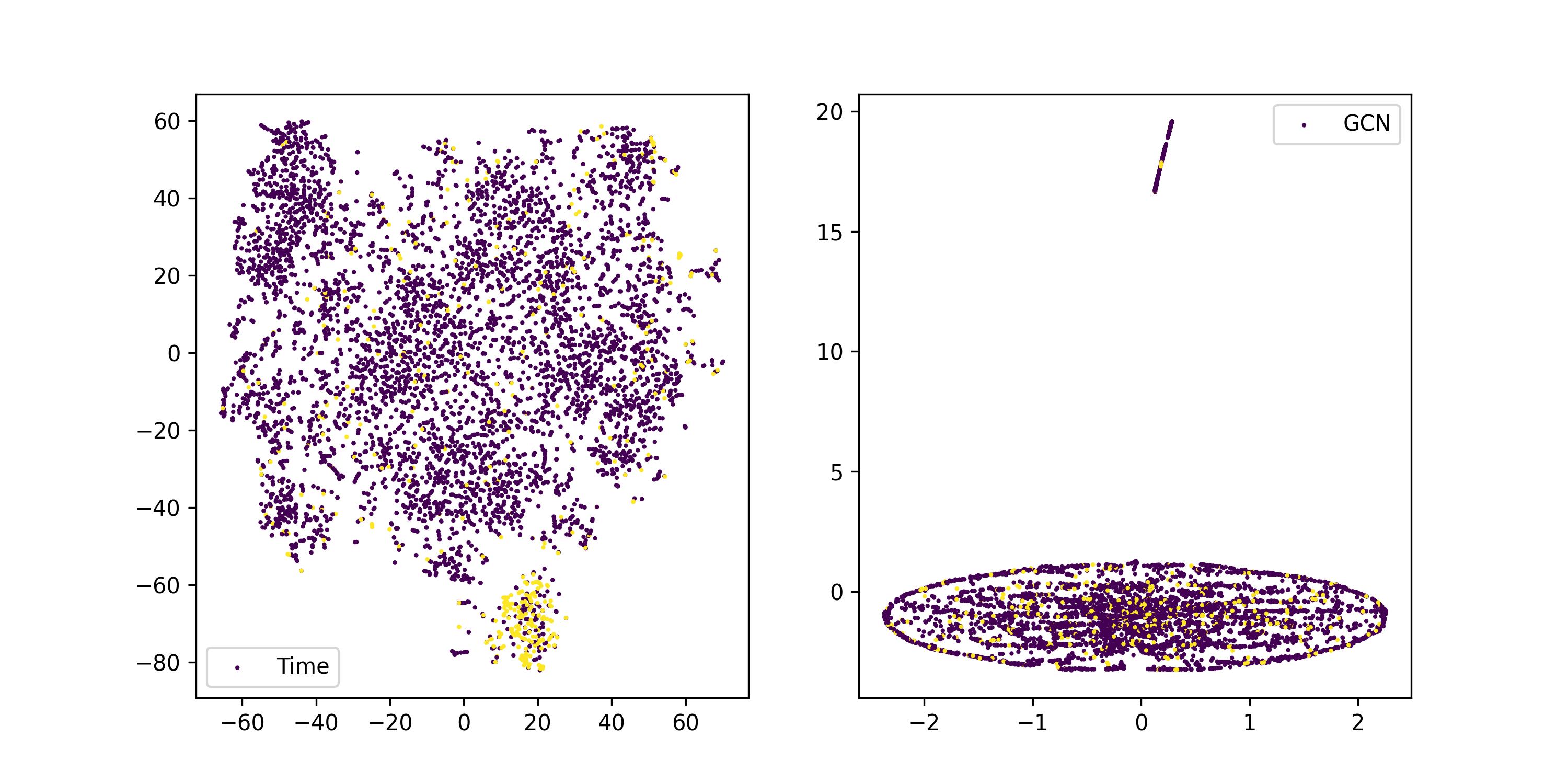}}
    \subfigure[Timestamp 10]{
    \includegraphics[width=2.8in]{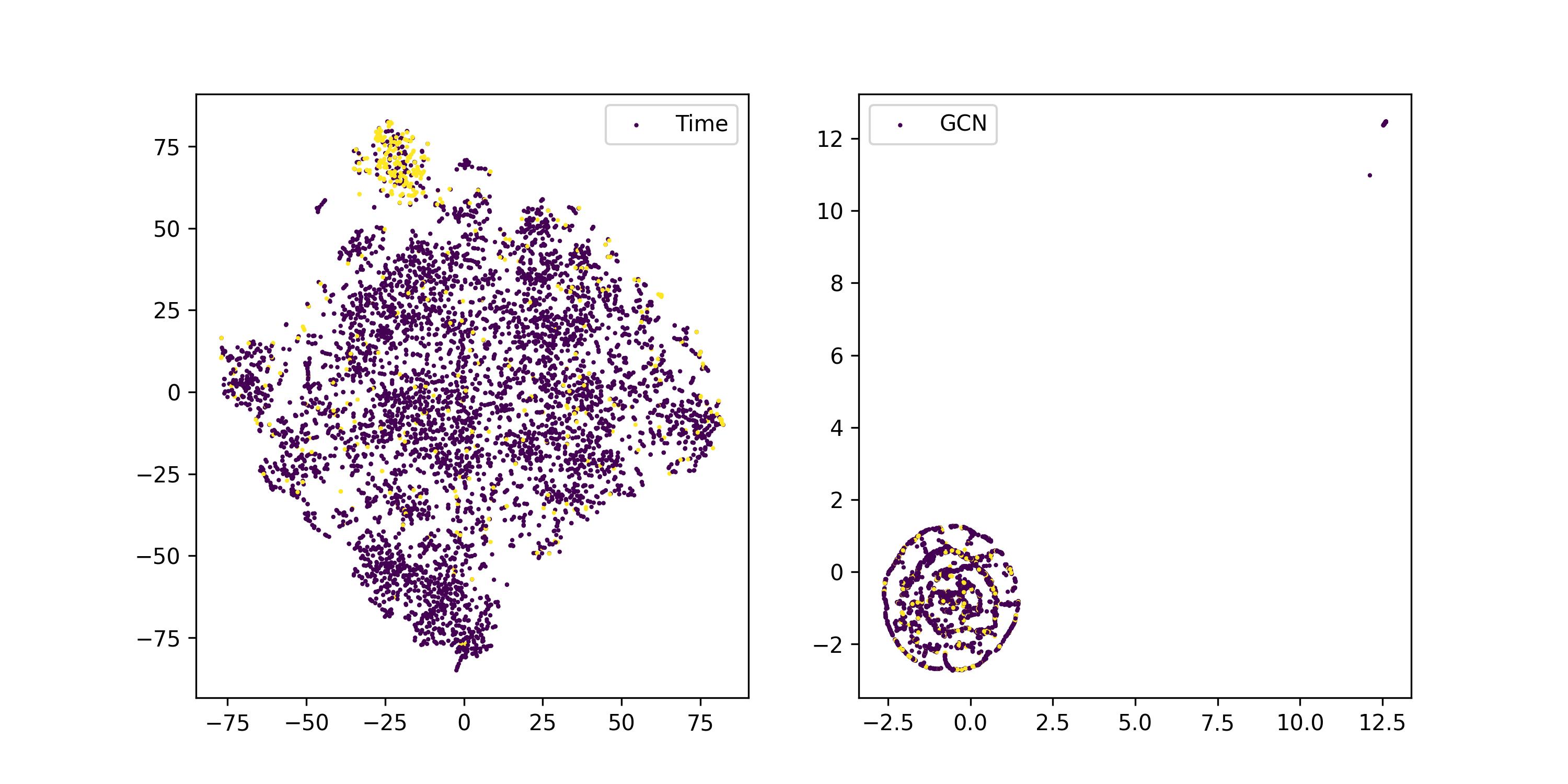}}
    \caption{Dynamic node embedding in bitcoin-alpha dataset }
    \label{fig:otcvis}
    \end{figure}
\end{center}

Since the node embedding learned from the model plays a key role in the downstream tasks, we assess the quality of learned dynamic node embedding by visualizing them in the plane. A node is labeled fraud (yellow) if it connects to more negatively rated edges than positively rated edges. In Figure 5 and Figure 6, the node embeddings are visualized using t-SNE \cite{van2008visualizing} technique. Figure \ref{fig:alphavis} shows the static node embedding generated from GCN and dynamic node embedding generated from MGS-TGCN for bitcoin-alpha dataset. Similarly, Figure \ref{fig:otcvis} shows the node embedding for bitcoin-otc dataset. 
\\
\\
Figure 5 and figure 6 are direct illustration of how the embeddings are distributed in the space. From the graphs, we have two observations:
\begin{itemize}
    \item The fraud nodes have more obvious clustering appearance in MGS-TGCN compared with static GCN. This explains why MGS-TGCN performs better in the edge classification task.
    \item The distribution of embeddings in MGS-TGCN is more stable as time evolves. The stableness is important in respect of explainability for practical applications. For example, the detected fraud communities in a consecutive of days are not supposed to vary too much.
\end{itemize}

\section{Conclusion and future work}
In this study, we propose a temporal GCN model MGS-TGCN with the training process guided by the motif and group balance information. To our knowledge, we are the first to simultaneously consider the temporal, local structural as well as the group balance information for evolving networks. Experimental results on bitcoin-alpha and bitcoin-otc datasets illustrates that MGS-TGCN outperforms baseline models by a wide margin.
\\
\\
Still, there are remaining works left for future development.
\begin{itemize}
    \item In this work, we consider networks in discrete-time setting, which overlooks the continuous-time information. 
    \item The motif matrices used in this study are computed from static graphs. Temporal motifs could better capture the local structural information.
    \item MGS-TGCN is a transductive learning method. An inductive learning approach is more practically useful for real-time decision making. In our future work, we will consider how to adapt MGS-TGCN for real-time inductive inference.
\end{itemize}

\section*{Acknowledgement}
The work described in this paper was partially supported by grants from the Research Grants Council of the Hong Kong Special Administrative Region, China (Project No. CityU C1143-20G) and (Project No. CityU 11504320).

%
%
%
 \bibliographystyle{splncs04}
 \bibliography{ref}

\end{document}